\documentclass[utf8]{frontiersinFPHY_FAMS}
\usepackage{url,hyperref,microtype,subcaption}
\usepackage[onehalfspacing]{setspace}
 \usepackage{multirow,array}
\usepackage{adjustbox}
\usepackage[ruled]{algorithm2e}
\usepackage{algorithmic}
\usepackage{float}
\usepackage{colortbl}

\usepackage{booktabs}
\def\keyFont{\fontsize{8}{11}\helveticabold }
\def\firstAuthorLast{Sample {et~al.}} %use et al only if is more than 1 author
\def\Authors{Qian Guo\,$^{1,*}$,Peiyuan Chen\,$^{2}$}

\def\Address{$^{1}$School of Economics and Management, Anhui Normal University, Wuhu, 241000, China \\
$^{2}$Oregon State University, Corvallis, Oregon, 97333, US
}
\def\corrAuthor{Qian Guo}

\def\corrEmail{guoqian5527@ahnu.edu.cn}

\begin{document}
\onecolumn
\firstpage{1}

%标题
\title {Construction and optimization of health behavior prediction model for the elderly in smart elderly care} 

\author[\firstAuthorLast ]{\Authors} %This field will be automatically populated
\address{} %This field will be automatically populated
\correspondance{} %This field will be automatically populated

\extraAuth{}% If there are more than 1 corresponding author, comment this line and uncomment the next one.
%\extraAuth{corresponding Author2 \\ Laboratory X2, Institute X2, Department X2, Organization X2, Street X2, City X2 , State XX2 (only USA, Canada and Australia), Zip Code2, X2 Country X2, email2@uni2.edu}

\maketitle

\begin{abstract}

%%% Leave the Abstract empty if your article does not require one, please see the Summary Table for full details.
\section{}%摘要部分

With the intensification of global aging, health management of the elderly has become a focus of social attention. This study designs and implements a smart elderly care service model to address issues such as data diversity, health status complexity, long-term dependence and data loss, sudden changes in behavior, and data privacy in the prediction of health behaviors of the elderly. The model achieves accurate prediction and dynamic management of health behaviors of the elderly through modules such as multimodal data fusion, data loss processing, nonlinear prediction, emergency detection, and privacy protection. In the experimental design, based on multi-source data sets and market research results, the model demonstrates excellent performance in health behavior prediction, emergency detection, and personalized services. The experimental results show that the model can effectively improve the accuracy and robustness of health behavior prediction and meet the actual application needs in the field of smart elderly care. In the future, with the integration of more data and further optimization of technology, the model will provide more powerful technical support for smart elderly care services.

\tiny
%关键词
 \keyFont{ \section{Keywords:}Smart elderly care, Health behavior prediction, Data privacy, Aging, Medical data analysis}
\end{abstract}

\section{Introduction}%引言部分
As the global population ages, the health issues of the elderly population are receiving increasing attention ~\cite{MA2023101808, FRISHAMMAR2023122319, ERNST2023107231}. According to statistics, the proportion of the global elderly population continues to rise. By 2050, the population aged 60 and above will account for 22\% of the worldwide population. This trend brings many challenges. In particular, in the context of limited medical resources and insufficient elderly care services, how to predict the health behaviors of the elderly through scientific and effective methods and take preventive measures promptly has become an urgent problem to be solved in many fields such as public health, social security, and smart elderly care. The prediction of the health behaviors of the elderly is not only related to improving individual health and quality of life but also has important significance for reducing medical costs and alleviating family and social burdens.

The research on the prediction of health behaviors of the elderly can be traced back to decades ago. Early studies mainly relied on traditional health assessment methods, such as regular physical examinations, questionnaires, and expert interviews. Although these methods can provide an overview of health status to a certain extent, their drawbacks are also becoming increasingly apparent. First, these methods are highly dependent on the participation and cooperation of the elderly themselves. The data acquisition process is complex and time-sensitive, making it difficult to reflect the dynamic health status of the elderly promptly. Second, traditional methods make it difficult to quantify individual differences and cannot meet the needs of personalized health management. With the rapid growth of the elderly population, traditional health prediction methods have gradually exposed problems such as insufficient data and low prediction accuracy. New technical means are urgently needed to improve the prediction effect.

In recent years, the rapid development of the Internet of Things (IoT), big data, and artificial intelligence technologies~\cite{Wang2024Theoretical,wang2021machine,yuan2025gta,li2024optimizing,wang2024recording,li2024deep,zhuang2020music,sui2024application,chen2024enhancing,wan2024image} has provided new opportunities for predicting the health behavior of the elderly. Smart elderly care has gradually become a frontier research field. By combining sensor networks, wearable devices, smart home systems, and artificial intelligence algorithms~\cite{cao2018expected,xi2024enhancing,de2023performance,chen2024mix,weng2024fortifying,yan2024application,gong2020research,peng2024automatic,wang2022classification,wang2024using}, real-time monitoring and analysis of the daily behavior and health data of the elderly can be achieved ~\cite{kulurkar2023ai, wang2024heart, addae2024smart}. Smart elderly care technology can not only help the elderly live independently but also effectively reduce the incidence of emergencies caused by health problems. For example, smart bracelets can monitor the elderly's heart rate, blood pressure, body temperature and other physiological data in real-time, and transmit the data to the cloud through wireless networks, and combine big data analysis technology to dynamically evaluate health behaviors. At the same time, smart home devices can identify potential health risks and provide timely warnings by monitoring the elderly's daily activities, such as getting up, going to bed, eating, and exercising. These technical means based on the Internet of Things and artificial intelligence have opened up new paths for research on predicting the health behavior of the elderly. However, although current smart elderly care technologies have shown great potential in elderly health management, they still face many challenges. First, the prediction of elderly health behaviors involves the processing of multi-source heterogeneous data, such as physiological data from wearable devices, activity data from smart homes, and electronic medical records and medication records ~\cite{kulurkar2023ai, hosseinzadeh2023elderly}. How to effectively integrate these data to form a high-quality prediction model is an urgent problem to be solved. Secondly, the health behaviors of the elderly are highly individualized and uncertain. There are significant differences in the living habits, health conditions, and disease risks of different individuals, which makes health behavior prediction face complex modeling challenges. In addition, how to provide accurate and timely health behavior predictions while ensuring data privacy and security is also an issue that needs to be considered in current smart elderly care research.

To cope with the multiple challenges in predicting the health behavior of the elderly, this paper proposes and implements a health behavior prediction platform for the elderly that integrates multi-source data. The platform collects the health data of the elderly in real-time through IoT devices and uses multimodal data fusion technology to effectively integrate data from different sources. The platform can comprehensively monitor and analyze the physiological parameters, daily activities, environmental factors, etc. of the elderly, and handle the problem of missing data through data interpolation and robustness enhancement algorithms~\cite{zheng2024identification,qiao2024robust,weng2024big,zheng2024triz,ren2024iot,huang2024risk,wang2018performance,wang2024deep,yang2019regional,wang2024cross,liu2024dsem}. The platform also pays special attention to data privacy protection, and adopts federated learning and differential privacy mechanisms to ensure that the data of the elderly are processed in a safe environment. Through these technologies, the platform can not only automatically collect and analyze data, but also provide personalized health management suggestions based on the health status of the elderly, thereby improving the quality of life of the elderly and reducing the care burden of society and family.

To verify the effectiveness of the proposed platform, this paper designed and conducted several experiments, which were tested in different scenarios such as community care, home care, and hospital care. The experimental data came from multiple sources, including physiological data from wearable devices of the elderly, activity data from smart home systems, and electronic medical record data from hospitals. The experimental results show that the proposed prediction platform has good adaptability and stability in different scenarios, and the prediction accuracy is significantly better than traditional methods. In the community care scenario, the platform can accurately predict the risk of falls of the elderly and issue early warnings in time; in the home care scenario, the platform can identify abnormal living habits of the elderly, such as long-term sedentary or increased nighttime activities; in the hospital care scenario, the platform predicts the risk of disease recurrence by analyzing the historical medical records of the elderly, and provides scientific intervention suggestions for medical staff.

The main contributions of this paper are as follows:
\begin{enumerate}
    \item Firstly, a health behavior prediction platform for the elderly that integrates multi-source data was designed and implemented, and effective data fusion and modeling methods were proposed, which can accurately predict the health behavior of the elderly and issue early warnings promptly.
    \item Secondly, the deep learning model proposed in this paper combines the advantages of convolutional neural networks and long short-term memory networks and enhances the capture of key health characteristics through the attention mechanism, thereby significantly improving the accuracy of prediction.
    \item Thirdly, through experimental verification in different elderly care scenarios, the wide applicability and robustness of the platform were demonstrated, providing theoretical and technical support for further research in the field of smart elderly care.
    \item Finally, this paper discusses the challenges and development directions in future research on predicting health behaviors of the elderly, especially proposing new ideas in multi-source data fusion, personalized health management, and data privacy protection.
    
\end{enumerate}

In short, with the rapid growth of the elderly population, how to use advanced technical means to predict and manage the health behavior of the elderly is an important issue that needs to be solved urgently. This paper demonstrates the broad prospects of the field of smart elderly care by building a health behavior prediction platform for the elderly that integrates multi-source data, and provides strong theoretical support and practical reference for future research.

\section{Related Work}
As an important part of smart elderly care, health behavior prediction for the elderly has received extensive attention and in-depth research in recent years. The progress of related work is mainly concentrated in the following aspects: improvement of traditional health behavior prediction methods, health monitoring systems based on the Internet of Things, and health behavior prediction in smart elderly care.

\subsection{Traditional health behavior prediction methods}

In the early research on predicting the health behavior of the elderly, traditional methods ~\cite{faul2023epigenetic, younis2024health, ibrahim2023older} mainly relied on questionnaires, health examinations, and expert evaluations. These methods collect basic health data of the elderly, such as blood pressure, blood sugar, heart rate, etc., and combine their living habits and medical history to assess health risks and predict possible health problems. However, these traditional methods face many challenges.
First, these methods are highly dependent on the participation and cooperation of the elderly themselves. The data collection process is usually complicated and the amount of data is limited. This static data cannot fully reflect the dynamic changes in the health status of the elderly, resulting in insufficient timeliness and accuracy of the prediction~\cite{li2024learning,jiang2024trajectorytrackingusingfrenet,liu2024td3basedcollisionfree,shen2024deep}. Second, traditional methods make it difficult to quantify individual differences, especially when facing a diverse elderly population, and it is difficult to provide personalized health management plans. In addition, methods such as health examinations and questionnaires often require high human and material resources are not sustainable, and cannot be promoted on a large scale.
Nevertheless, traditional methods laid the foundation for early research on the health of the elderly and provided a reference for subsequent research. With the advancement of data collection technology and analysis methods, researchers have gradually realized the need to combine emerging technologies to make up for the shortcomings of traditional methods, to achieve more accurate and real-time health behavior prediction.

\subsection{Health monitoring system based on internet of things}
The development of IoT technology~\cite{zhou2024adapi} has brought new opportunities for monitoring and predicting the health behavior of the elderly. In recent years, researchers have increasingly used IoT devices for real-time health monitoring, such as smart bracelets, smart homes, and wearable sensors. These devices can collect physiological data and daily activity data of the elderly in real-time, and transmit the data to the cloud for analysis through wireless networks, thereby realizing dynamic monitoring of health status ~\cite{facchinetti2023can, sorwar2023factors, he2023new,peng2024maxk,xie2023accel}.
Health monitoring systems based on IoT have many advantages. First, these systems can collect data in real-time and discover changing trends in the health status of the elderly through data analysis. For example, smart bracelets can monitor data such as heart rate, blood pressure, and body temperature of the elderly in real-time and immediately issue an alarm when an abnormality is detected. Similarly, smart home systems can identify potential health risks by monitoring the daily activities of the elderly, such as getting up, eating, and going out. Second, the data collection process of IoT devices~\cite{richardson2024reinforcement,zhu2024complex,xu2022dpmpc,weng2024comprehensive,jiang2020dualvd} is usually automated, reducing the reliance on the active participation of the elderly and ensuring the continuity and integrity of the data. In addition, the data collected by IoT devices can be combined with other data sources (such as electronic medical records, medication records, etc.) to achieve more comprehensive health behavior prediction ~\cite{hosseinzadeh2023elderly, kulurkar2023ai}.
However, health monitoring systems based on IoT also face some challenges. For example, how to effectively process and analyze a large amount of heterogeneous data is a key issue. Different types of IoT devices may generate data in different formats. The fusion and analysis of these data require powerful data processing capabilities and advanced algorithm support. In addition, data privacy and security are also issues that cannot be ignored. How to achieve data sharing and analysis while ensuring data security and user privacy is one of the important directions of current research.

\subsection{Prediction of health behavior in smart elderly care}
Smart elderly care is a new elderly care model that combines information technology, artificial intelligence, and elderly care services. Its core goal is to improve the quality of life and safety of the elderly through technical means. In the framework of smart elderly care, health behavior prediction plays a vital role. Through real-time analysis and prediction of daily behavior data of the elderly, the smart elderly care system can provide personalized health management services for the elderly, reduce health risks, and improve the quality of life.
Health behavior prediction in smart elderly care covers multiple fields, including fall prediction, chronic disease management, emotional state monitoring, etc. ~\cite{yu2024motion, bargiotas2023preventing}. For example, falls are a common and serious health problem for the elderly. Using machine learning and deep learning technologies, researchers have developed a variety of fall detection and prediction systems. These systems can identify the risk of falls in advance by analyzing the gait, posture, and movement patterns of the elderly, and issue alarms to reduce the incidence of accidents. For example, based on data from wearable devices, researchers can analyze the gait of the elderly through convolutional neural networks (CNN) and long short-term memory networks (LSTM) models to predict potential fall risks~\cite{RAN2024103664,HAO2024102551}.

In addition, chronic disease management is also an important application area in smart elderly care. By monitoring the daily activities and physiological data of the elderly, the smart elderly care system can promptly identify early signs of chronic disease onset and provide personalized intervention measures. For example, using IoT devices and big data analysis technology, researchers can monitor indicators such as blood sugar levels and heart rate changes of the elderly, and combine historical data to predict risks, thereby helping medical personnel to develop more accurate treatment plans.
Emotional state monitoring is also an emerging direction in smart elderly care in recent years. The mental health of the elderly is also crucial, and fluctuations in emotional state may have a significant impact on their overall health. By analyzing the elderly’s voice, facial expressions, and social activity data, the smart elderly care system can assess their emotional state and provide psychological counseling and support when abnormalities are detected. These technical means based on artificial intelligence and data analysis provide more comprehensive and personalized health management services for the elderly ~\cite{oyebode2023machine, calbimonte2023decentralized, illert2023german}.

In summary, as an important part of smart elderly care, the prediction of the health behavior of the elderly has made significant research progress. Traditional health behavior prediction methods have laid the foundation for research, but they are insufficient in the face of dynamic changes and personalized needs. The health monitoring system based on the Internet of Things provides a real-time and comprehensive means of data collection, but how to effectively process and analyze this data remains a challenge. Health behavior prediction in smart elderly care covers multiple application areas, and through advanced technical means, it provides personalized health management services for the elderly. Future research will further explore the application of multi-source data fusion technology to improve the accuracy and reliability of health behavior prediction, and solve problems such as data privacy and security, to provide a more comprehensive solution for the health management of the elderly.

\section{Methodology}

\subsection{Related theories}
In the design and implementation of the smart elderly care platform, the selection and application of theoretical foundations are crucial. To ensure that the platform can effectively meet the needs of the elderly, this study combines theories from multiple fields such as geriatrics, health behavior, Internet of Things technology, big data analysis, and artificial intelligence ~\cite{lin2023prediction, das2023application} and applies these theories to the design, data processing, and decision support of the platform.

The elderly population has unique needs in terms of physiological, psychological and social characteristics. Physiologically, the elderly's body functions gradually decline, they are prone to chronic diseases, and their ability to adapt to the external environment decreases. Psychologically, the elderly are prone to loneliness and anxiety, and their emotional needs are more prominent. Socially, the elderly's social circle is relatively small, and they rely on the support system of family and community. Therefore, when designing a smart elderly care platform, these characteristics of the elderly population must be fully considered to ensure that the platform can provide services that meet actual needs.
To cope with these characteristics of the elderly population, the platform design is based on the Healthy Aging theory ~\cite{peel2005behavioral}. This theory emphasizes delaying the process of health decline and improving the quality of life of the elderly through preventive health management, personalized care, and active social participation. Specifically in the implementation of the platform, the Healthy Aging theory guides us to design functional modules such as real-time health monitoring, psychological care, and social support, to provide comprehensive health management services for the elderly.

Health behavior theory provides theoretical support for the prediction of health behavior of the elderly. Health behavior research shows that health behavior is affected by multiple factors such as personal physiology, psychology, social environment, and behavioral habits. These factors interact with each other and jointly determine the occurrence and change of health behavior. Taking the Theory of Planned Behavior (TPB) ~\cite{ajzen1991theory} as an example, this theory believes that an individual's behavioral intention is influenced by attitudes, subjective norms, and perceived behavioral control, which are the key to behavior change.
In the smart elderly care platform, health behavior theory is used to guide the design of behavior prediction models ~\cite{glanz2015health, painter2008use}. By collecting physiological data, activity data, and psychological state data of the elderly, the platform can comprehensively evaluate the influencing factors of their health behavior and predict possible health risks. The platform uses health behavior theory to integrate these factors into the prediction model, thereby improving the accuracy and effectiveness of the prediction. For example, the platform can predict possible falls or abnormal health events of the elderly based on their behavioral intentions and daily activity patterns, and provide corresponding warnings and intervention suggestions.

The core of the smart elderly care platform lies in the collection and analysis of data, so the IoT and big data theory occupy an important position in the construction of the platform. The Internet of Things theory supports the collection and transmission of health behavior data of the elderly. Through smart devices and sensor networks, the platform can monitor the physiological indicators, behavior patterns, and environmental conditions of the elderly in real time. The data is transmitted to the cloud or edge computing devices through the Internet of Things system for processing, realizing dynamic analysis, and real-time feedback of health data.
Big data theory provides a basis for the storage, processing, and analysis of data. The platform's data sources are extensive, including real-time data from smart devices, historical health records, and social and psychological status data. Through big data technology, the platform can effectively process these massive and diverse data and extract high-value health behavior characteristics. At the same time, based on big data analysis theory, the platform can identify complex associations hidden in the data and generate personalized health predictions and management plans.

In addition, the platform also combines expert systems and decision support theories and uses AI algorithms to provide scientific health management suggestions for the elderly and caregivers. By learning the health patterns of the elderly population, the platform automatically generates personalized care plans and provides emotional care and social support. This AI-based decision support system can not only improve the accuracy of health management but also significantly enhance the user experience of the platform.

\begin{figure}
    \centering
    \includegraphics[width=1\linewidth]{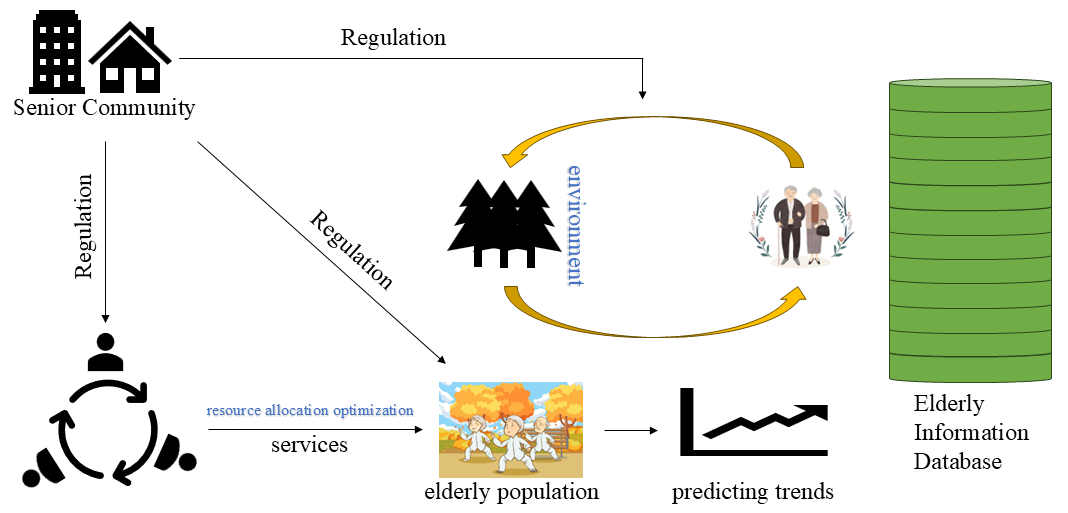}
    \caption{The overall framework structure of our model for predicting health behavior among the elderly.}
    \label{model}
\end{figure}

\subsection{Smart elderly care service model}
The core of the smart elderly care service model is to achieve all-around support for the elderly population through effective resource allocation, accurate health behavior prediction, and reasonable cost-benefit analysis. To build this model, we combined the theoretical foundations of multiple disciplines and described them through mathematical formulas to ensure that the model is operational in practice and rigorous in theory. The overall framework is shown in Figure \ref{model}.
The main goal of smart elderly care is to provide comprehensive health management services for the elderly through information technology and intelligent means, including daily health monitoring, emergency warnings, chronic disease management, and quality of life improvement. In this context, our model first starts from the three core perspectives of resource allocation, dynamic changes in health behavior, and cost-benefit analysis to ensure that the system can effectively respond to the diverse needs of the elderly population.

\subsubsection{Resource allocation and optimization}
In smart elderly care services, the optimal allocation of resources is the key to maximizing the effect of health management. Resource allocation issues usually involve multiple participants, such as the government, social capital, medical institutions, and elderly families. When resources are limited, how to allocate resources to maximize health benefits is the primary issue in model design.
Assume that the allocation of resources $R_i$ follows the following utility function:
\begin{equation}
    U_i(R_i)=\frac{a_i\cdot R_i^{\theta_i}}{1+b_i\cdot R_i^{\delta_i}}
\end{equation}
among them, $a_i$ and $b_i$ are utility coefficients, $\theta_i$ and $\delta_i$ are utility indices. This utility function reflects the phenomenon of diminishing marginal returns of resource allocation, that is, as resource input increases, the increase in utility gradually decreases. When resource allocation reaches a certain level, the effect of continuing to increase resource input will become insignificant. This feature is particularly important in smart elderly care, because resources (such as funds, medical equipment, nursing staff, etc.) are usually limited, so they need to be reasonably allocated to maximize the overall benefit.
In practical applications, the solution of the utility function can be achieved through optimization algorithms such as the Lagrange multiplier method. By solving the maximum value of the utility function, the system can determine the optimal resource allocation strategy for each participant. This strategy can be dynamically adjusted according to variables such as the health status, social support, and environmental factors of the elderly to ensure the effective use of resources.

\subsubsection{Dynamic changes in health behaviors}
The health status of the elderly is not only closely related to resource input but also affected by their daily behavior, psychological state, social support, and external environment. To model these complex factors, we introduced the dynamic change equation of health behavior. Assume that the change of the health status $H(t)$ of the elderly over time $t$ can be described by the following differential equation:
\begin{equation}
    \frac{dH(t)}{dt}=\alpha_1\cdot S(t)+\alpha_2\cdot P(t)+\alpha_3\cdot E(t)-\beta\cdot C(t)
\end{equation}
in this equation, $S(t)$ represents the intensity of social support, $P(t)$ represents the psychological state, $E(t)$ represents the influence of environmental factors, and $C(t)$ represents the cost of health behavior. Parameters $\alpha_1$, $\alpha_2$, $\alpha_3$ represent the positive effects of different factors on health status, and $\beta$ represents the negative effect of cost on health status.
This equation can be numerically solved to obtain the trajectory of changes in the health status of the elderly. This model is particularly suitable for predicting the evolution of long-term health behaviors, such as the cumulative impact of long-term exercise and healthy diet on the health status of the elderly. Through this model, the smart elderly care service platform can provide personalized health management advice for the elderly and take intervention measures at the early stage of health deterioration.

\subsubsection{Interactions between environmental factors and health status}
To further characterize the complex relationship between the health behavior of the elderly and the external environment, we introduce external environmental factors into the dynamic change equation of health behavior. Assuming that the health state of the elderly $H(t)$ is affected by the external environment $E(t)$, this can be described by the following equation:
\begin{equation}
    \frac{dH(t)}{dt}=\alpha\cdot\frac{R_h(t)}{1+\kappa E(t)}-\beta\cdot C(t)
\end{equation}
where, $R_h(t)$ represents the resources obtained by the elderly, and $\kappa$ is the inhibition coefficient of the environment on health status. The equation shows that as the environment deteriorates (such as air pollution, noise pollution climate change, etc.), the improvement of the health status of the elderly will be inhibited. This means that in harsh environments, more resources are needed to maintain the health status of the elderly.
To alleviate this impact, the smart elderly care service platform can combine environmental monitoring data to dynamically adjust the health management strategy of the elderly population. For example, during periods of severe air pollution, the platform can advise the elderly to reduce outdoor activities and provide them with appropriate indoor exercise plans. At the same time, the platform can also monitor environmental data in real-time through smart devices and automatically adjust resource allocation according to environmental changes.

\subsubsection{Dynamic resource allocation and health management}
In the process of providing health management services, the smart elderly care platform needs to dynamically adjust the resource allocation strategy according to the changes in the health status of the elderly. Assume that resource allocation follows the following rules:
\begin{equation}
    R_h(t)=\frac{\lambda\cdot\exp(\gamma H(t))}{1+\exp(\gamma H(t))}
\end{equation}
among them, $\lambda$ is the total amount of resources, and $\gamma$ represents the sensitivity coefficient of health status. The equation shows that when the health status of the elderly is poor (that is, $H(t)$ is low), resource allocation will increase significantly to improve the health status; when the health status is good, the increase in resource allocation will tend to be flat. This dynamic adjustment mechanism enables smart elderly care services to provide the most appropriate resource support in different periods and avoid the problem of resource waste or insufficient allocation.
In addition, we also consider the impact of social capital on the health behavior cost of the elderly. Assuming that the participation of social capital indirectly affects the health status of the elderly by affecting the health behavior cost $C(t)$, we can describe this process through the following equation:
\begin{equation}
    C(t)=C_0\cdot\left(1-\frac{\delta S_c(t)}{1+\eta R_m(t)}\right)
\end{equation}
among them, $C_0$ is the initial health behavior cost, $S_c(t)$ is the support of social capital for the health behavior of the elderly, $R_m(t)$ is the resource input of social capital, $\delta$ and $\eta$ represent the impact factor and resource sensitivity coefficient of social capital respectively. This formula shows that with the increase of social capital support and investment, the health behavior cost of the elderly will decrease accordingly, thereby promoting the improvement of health status.

\subsubsection{Cost-benefit analysis and optimization}
In smart elderly care services, cost-benefit analysis is the key to achieving rational resource allocation. Assume that the total benefit $E(T)$ of smart elderly care services can be decomposed into the weighted sum of health benefits, social benefits, and economic benefits:
\begin{equation}
    E(T)=\gamma_1\cdot E_h(T)+\gamma_2\cdot E_s(T)+\gamma_3\cdot E_e(T)
\end{equation}
At the same time, the total cost of the system $B(T)$ can be decomposed into the weighted sum of direct cost, operating cost, and maintenance cost:
\begin{equation}
    B(T)=\lambda_1\cdot C_d+\lambda_2\cdot C_o+\lambda_3\cdot C_m
\end{equation}
Therefore, the cost-benefit ratio (CBR) can be expressed as:
\begin{equation}
    \mathrm{CBR}=\frac{E(T)}{B(T)}
\end{equation}

By analyzing the cost-benefit ratio in different scenarios, the platform can dynamically adjust the resource allocation strategy to maximize benefits. For example, when resources are sufficient, the platform can prioritize health benefits; when resources are tight, the system can be kept running efficiently by optimizing operation and maintenance costs.

Through the above platform, we can comprehensively model resource allocation, health status evolution, and the role of social capital in the smart elderly care service model. Each formula represents the adjustment mechanism of the platform to deal with different scenarios during actual operation. Through these mathematical models, the platform can achieve dynamic management and personalized services for the elderly population, thereby effectively improving the efficiency and effectiveness of health management.
These models can be further extended to complex scenarios. For example, in a multi-player game model, different stakeholders (such as the government, social capital, and elderly families) may choose different strategies based on their respective goals. In this case, the allocation of resources and the sharing of costs can be coordinated through a multi-objective optimization model to find a balance between the interests of multiple parties.
In addition, these models can be calibrated with real-time data, allowing the smart elderly care platform to adaptively adjust parameters to meet the health management needs of the elderly in different periods and under different environmental conditions. For example, during periods of climate change or infectious disease epidemics, the platform can increase resource allocation and adjust the parameters of health behavior prediction models in real-time, thereby reducing health risks among the elderly population.

\section{Experiment}

\subsection{Difficulties in predicting health behaviors of the elderly}
In the process of building a smart elderly care service platform, the prediction of the health behavior of the elderly is a key link. However, due to the particularity of the elderly population and the complexity of their health conditions, health behavior prediction faces many challenges and difficulties ~\cite{lima2012self, bertini2018predicting, gong2016factors}. The following are several major difficulties.

\textbf{Data diversity and heterogeneity.} The health behaviors of the elderly are affected by many factors, including physiological, psychological, social, and environmental factors. To accurately predict health behaviors, the platform needs to integrate multi-source data, such as physiological data from wearable devices, activity data from smart home devices, medical records, mental health assessments, and data from environmental sensors. These data not only come from diverse sources, but also have different formats, frequencies, and accuracies, and are highly heterogeneous. How to effectively integrate these diverse data and eliminate noise and redundant information is a major challenge in prediction.

\textbf{Complexity and individual differences in health status.} The health status of the elderly is often complex and highly individual. Even in similar age groups, there may be significant differences in physical function, medical history, lifestyle, and coping mechanisms. In addition, as they age, the decline of physical function and the onset of diseases in the elderly are often irregular and unpredictable. Health behavior prediction models need to be able to capture these complexities and differences to accurately predict individual health behaviors. However, traditional prediction methods are often based on group statistical models, which make it difficult to accurately model individuals.

\textbf{Long-term dependency and data loss issues.} Predicting the health behavior of the elderly requires modeling and analysis based on long-term historical data. However, in practical applications, it is not easy to obtain long-term data. On the one hand, the health status of the elderly may be unstable or the improper use of technical equipment, resulting in discontinuous data records; on the other hand, the elderly may have incomplete data collection due to cognitive decline or low technology acceptance. In addition, the data of the elderly group may also be affected by multiple factors such as socioeconomic status, cultural background, and living environment, further increasing the complexity and missing problems of the data.

\textbf{The suddenness and nonlinearity of behavioral changes.} The health behavior and health status of the elderly may change suddenly, such as sudden falls, acute illness attacks, or sharp fluctuations in emotional states. This suddenness and nonlinearity make prediction more difficult. Traditional prediction methods usually rely on the stationary and linear assumptions of data, but they are often powerless in the face of the highly nonlinear and unpredictable health behavior of the elderly. Therefore, the prediction model needs to have strong flexibility and robustness to cope with these sudden changes.

\textbf{Data privacy and ethical issues.} Elderly health data usually involves sensitive personal information, such as health status, behavioral habits, emotional state, etc. How to protect the privacy and data security of the elderly during data collection, storage, and analysis is an important ethical issue. In addition, in data-driven prediction models, how to ensure the fairness of the algorithm and avoid discrimination or wrong decisions caused by data bias is also a challenge that needs to be solved urgently.
To cope with the above difficulties, this study comprehensively considered multi-source data fusion, personalized modeling, data missing processing, nonlinear prediction methods, and privacy protection mechanisms in the experimental design to build a model that can accurately and real-time predict the health behavior of the elderly. Next, the specific design of the experiment, the data processing process and the implementation of the model will be described in detail.

\subsection{Implementation details}
\textbf{Dynamic Data Collection.}
The proposed elderly health behavior prediction platform integrates real-time data collection using IoT devices, such as wearable health monitors, smart home sensors, and environmental sensors. These devices continuously capture dynamic health indicators (e.g., heart rate, blood pressure, movement patterns) and environmental factors (e.g., temperature, humidity, air quality). The data is transmitted in real-time to a central server via secure wireless communication protocols. This dynamic data collection enables the system to continuously monitor changes in health status, capturing both gradual trends and sudden fluctuations, thus supporting timely intervention.

\textbf{Feedback Mechanisms.}
The platform incorporates an adaptive feedback mechanism on multiple levels: user feedback, family and caregiver alerts, and healthcare provider notifications. When a potential health risk is detected, the system immediately alerts relevant stakeholders via mobile apps, SMS, or emails. For the elderly user, personalized feedback is delivered through a user-friendly interface, including daily health reports, reminders for medication or physical activity, and warnings about potential health risks. Caregivers and healthcare providers receive summaries of abnormal patterns, along with recommended interventions based on historical data and current health status. This continuous feedback loop ensures that all stakeholders are informed in real-time, enabling proactive management of the elderly's health.

\textbf{Control Algorithms.}
The platform employs advanced control algorithms that dynamically adjust health recommendations and alerts based on the individual’s ongoing health status and historical data. These control algorithms are built using rule-based systems combined with predictive models to ensure responsiveness and adaptability. For example, if a significant drop in activity levels is detected alongside a rise in heart rate, the control algorithm will flag this combination as a potential health risk and escalate the level of alerts. Moreover, the system periodically retrains these algorithms using new data to improve accuracy and relevance, adapting to the changing needs and conditions of the elderly individual. By incorporating both automated adjustments and clinician-configurable thresholds, the control algorithms provide a balance between user customization and system-driven optimization.

\textbf{Revised Evaluation Strategy.}
To effectively assess the platform's real-world applicability and effectiveness, the evaluation process includes extensive testing under real-life conditions. Continuous data from various elderly users is analyzed to assess the accuracy of health predictions, the timeliness of feedback, and the relevance of control adjustments. Additionally, stakeholder feedback is gathered regularly to refine the feedback mechanisms and control algorithms further. This real-world testing provides essential insights into the platform’s adaptability, scalability, and reliability, ensuring it can meet the demands of dynamic and complex elderly care environments.

\subsection{Model design and experimental results analysis}
To address various challenges in predicting the health behavior of the elderly and verify the effectiveness of our proposed method, this study designed and implemented a complete health behavior prediction service model for the elderly. The model aims to address challenges such as data diversity and heterogeneity, complexity and individual differences in health status, long-term dependence, and data loss, sudden and nonlinear behavior changes, and data privacy and ethical issues.

\subsubsection{Design of service model for the elderly}
\textbf{Data collection.} Considering the diversity and heterogeneity of data, the model first collects multi-source data in real-time through IoT devices (such as wearable devices, smart home sensors, etc.). These data include physiological parameters of the elderly (such as heart rate, blood pressure, body temperature, etc.), daily activities (such as steps, sleep patterns, etc.), psychological states (such as emotions, stress levels, etc.), environmental information (such as air quality, temperature and humidity, etc.), and medical records (such as medical history, medication information, etc.). To solve the problem of data heterogeneity, we use multimodal data fusion technology to standardize and fuse data from different sources to form a unified feature representation. Specifically, we use a fusion network based on the attention mechanism to extract key health behavior features by weighting and merging the importance of different data sources.

\textbf{Data missing processing and robustness enhancement module.} To address the long-term dependency and data missing issues, we integrated data interpolation and robustness enhancement algorithms into the model. For short-term data missing, we used an interpolation method based on Gaussian processes to infer missing values and complete the data. For long-term data missing, we used a self-supervised learning framework to improve the model's robustness to missing data by constructing comparative learning tasks, ensuring that prediction accuracy is not affected by data missing.

\textbf{Privacy protection and security module.} Considering data privacy and ethical issues, the model is designed with special attention to data security and privacy protection. We use federated learning technology to train data on local devices, avoid uploading data to the cloud, and reduce the risk of privacy leakage. In addition, through the differential privacy mechanism, we ensure that during the data analysis and model training process, even if the data is maliciously accessed, no specific personal information can be obtained.

\textbf{Market research analysis and data-driven decision-making.} Before model design, this study conducted detailed market research to understand the gap between current market demand and existing smart elderly care services and guided the model design in a data-driven manner. By analyzing market research data, model functions can be optimized to ensure that they meet user needs. Table \ref{tab1} summarizes the importance of different functions in market research and the comparison of their availability in the current market. As shown in Table 1, real-time monitoring and prediction accuracy are the functions that users care about most, but there is still much room for improvement in their availability in the market.

\begin{figure}
    \centering
    \includegraphics[width=1\linewidth]{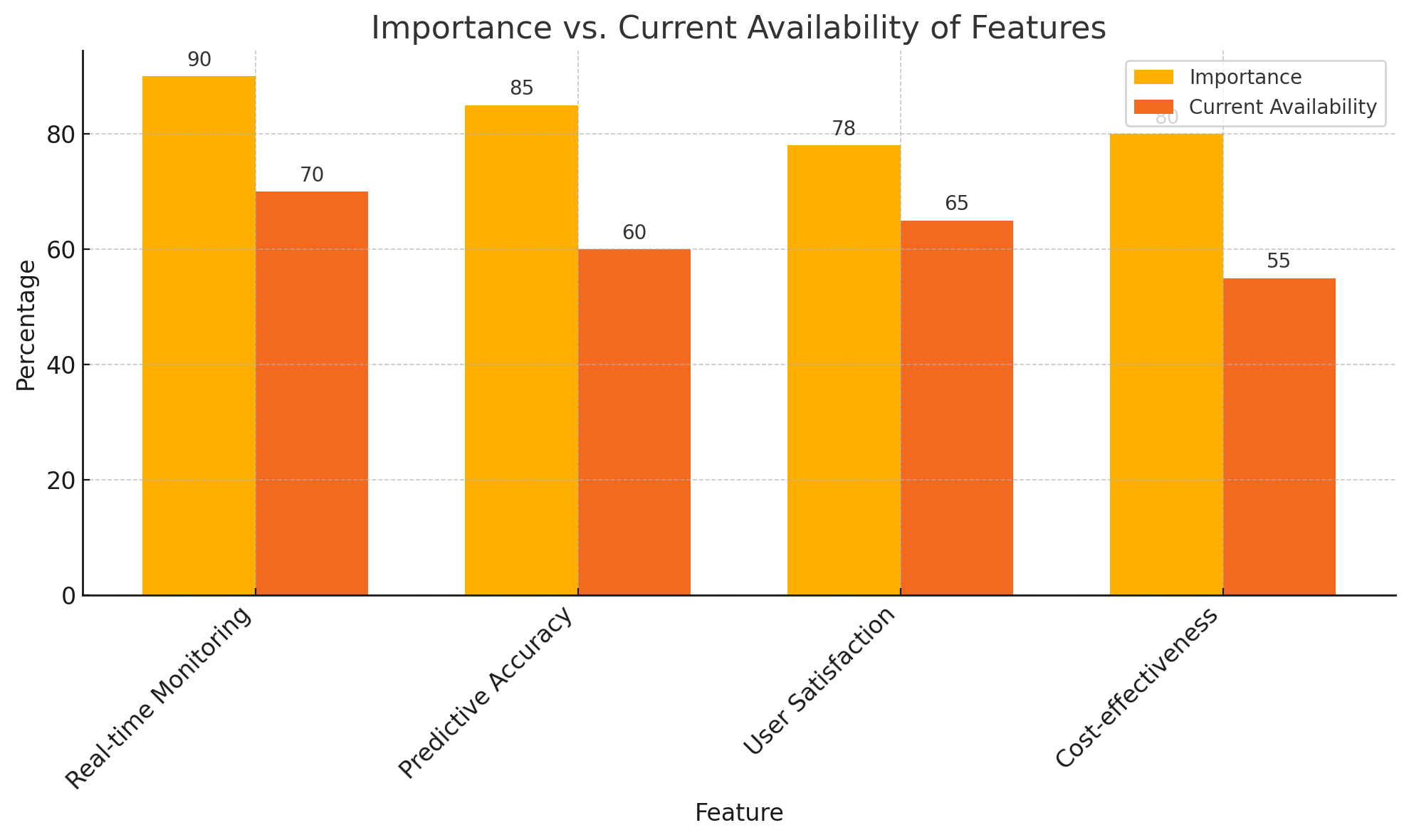}
    \caption{Feature importance versus current usability.}
    \label{fig1}
\end{figure}

\begin{figure}
    \centering
    \includegraphics[width=1\linewidth]{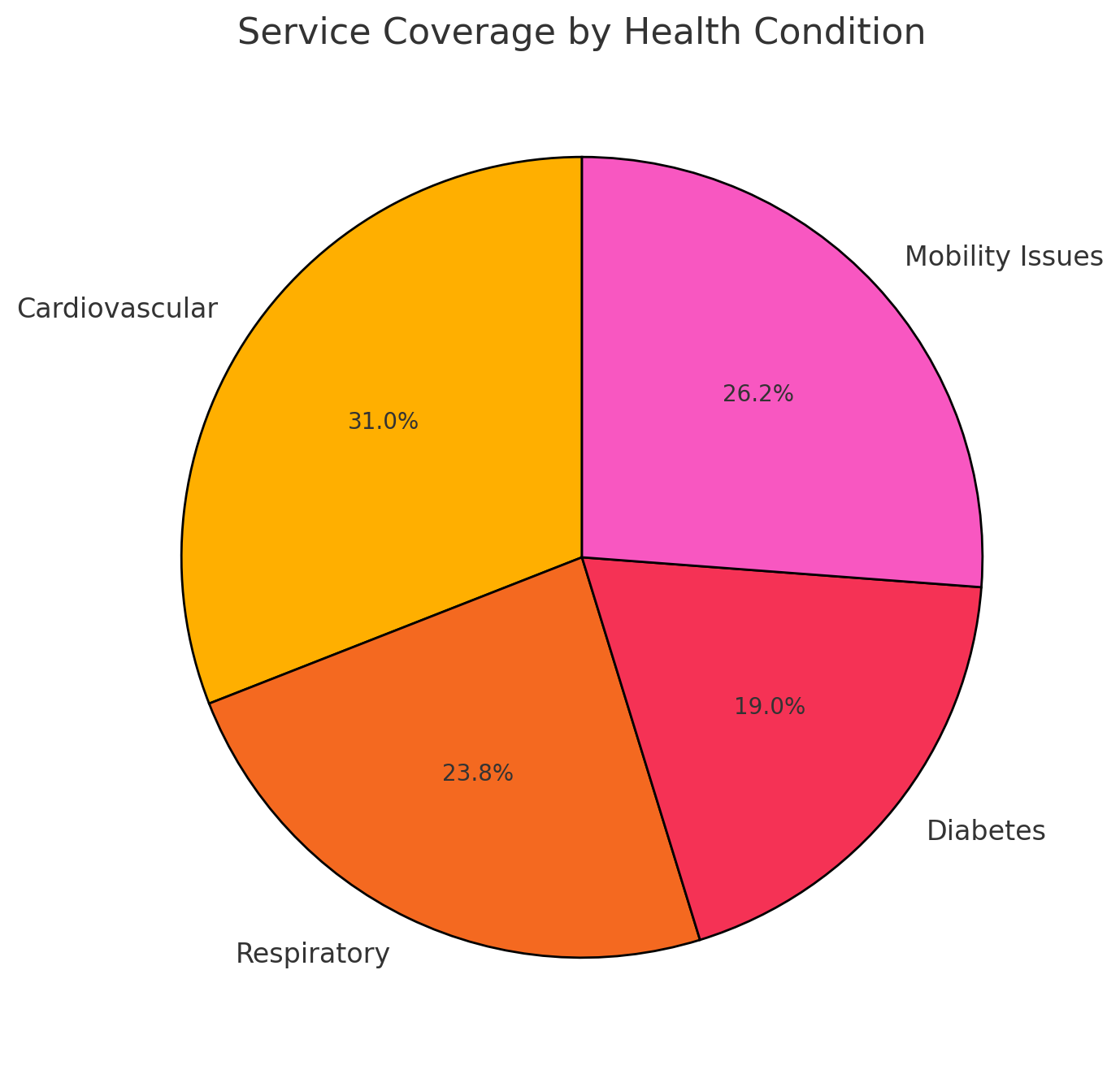}
    \caption{Analysis of health status service coverage.}
    \label{fig2}
\end{figure}

\begin{table}[htbp]
\centering
\caption{Comparison of Feature Importance, Availability, Technical Difficulty, and Future Coverage.}
\resizebox{\linewidth}{!}{
\begin{tabular}{|l|c|c|c|c|}
\hline
\textbf{Feature}           & \textbf{Importance (\%)} & \textbf{Current Availability (\%)} & \textbf{Technical Difficulty (1-5)} & \textbf{Expected Coverage in 5 Years (\%)} \\ \hline
Real-time Monitoring       & 90                       & 70                                & 4                                   & 95                                        \\ \hline
Predictive Accuracy        & 85                       & 60                                & 5                                   & 90                                        \\ \hline
User Satisfaction          & 78                       & 65                                & 3                                   & 85                                        \\ \hline
Cost-effectiveness         & 80                       & 55                                & 4                                   & 88                                        \\ \hline
\end{tabular}}
\label{tab1}
\end{table}

\begin{figure}
    \centering
    \includegraphics[width=1\linewidth]{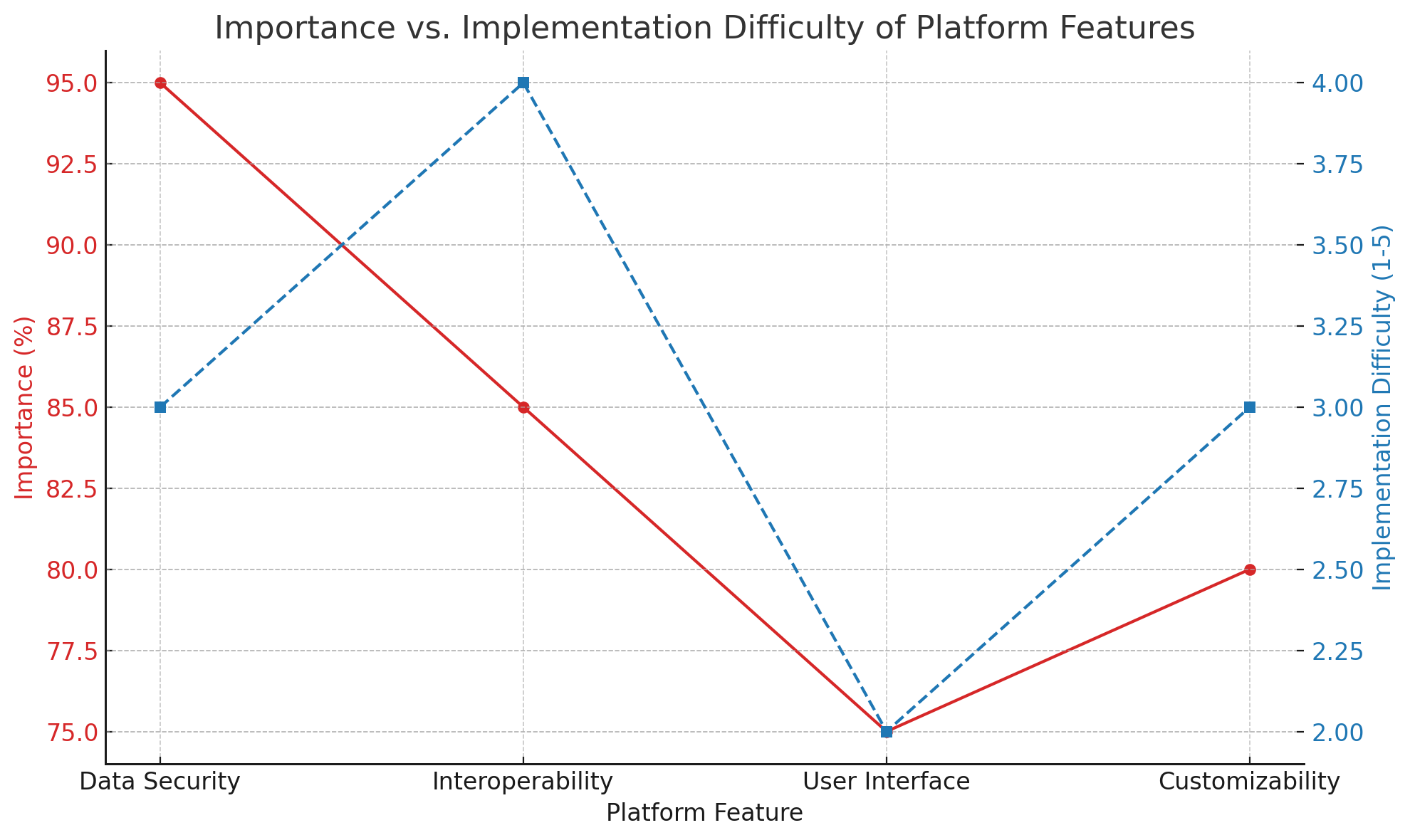}
    \caption{Analysis of the importance of platform functions and difficulty of implementation.}
    \label{fig3}
\end{figure}
\begin{figure}
    \centering
    \includegraphics[width=1\linewidth]{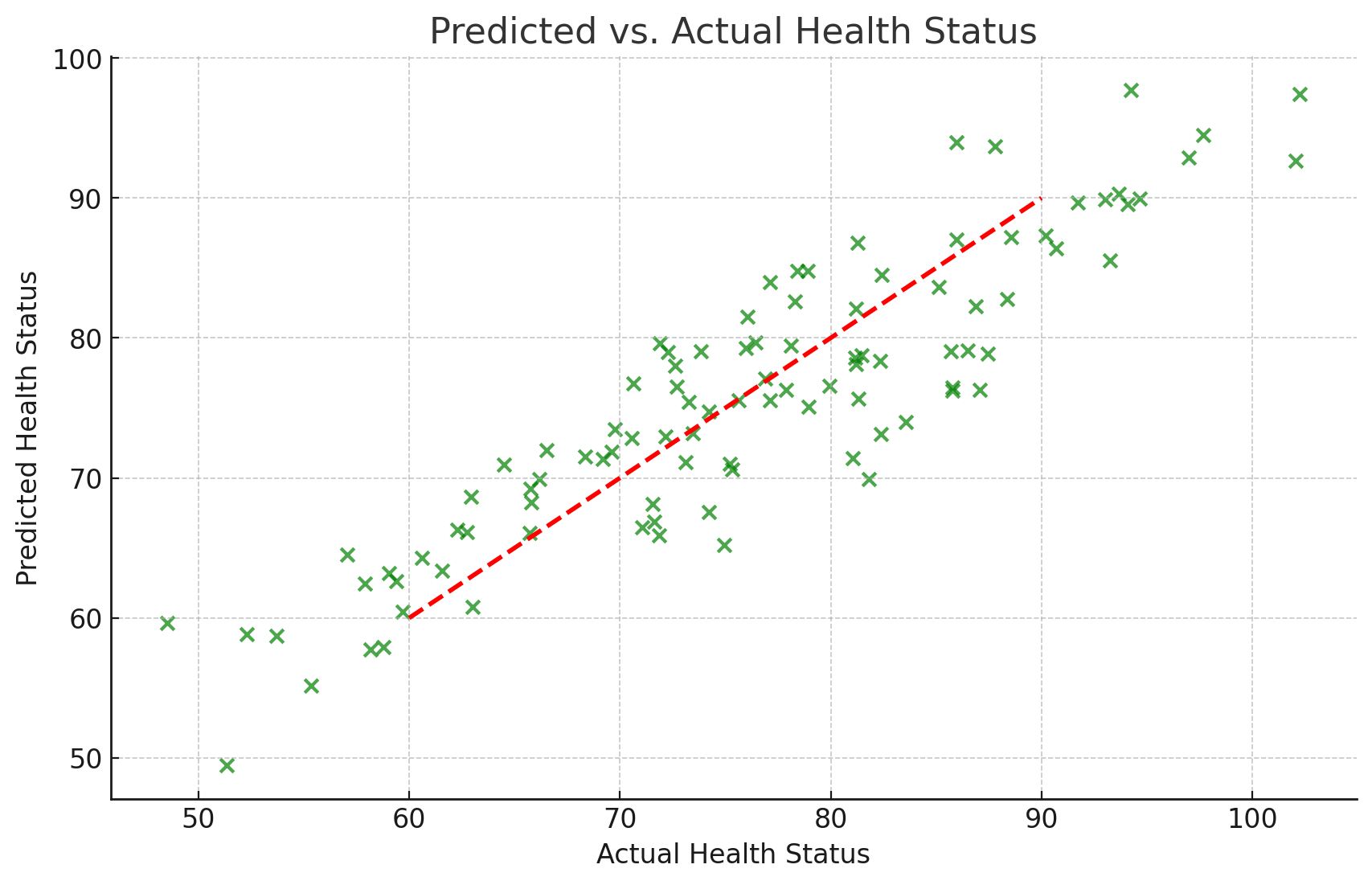}
    \caption{Comparison of predicted health status with actual health status.}
    \label{fig4}
\end{figure}

\begin{figure}
    \centering
    \includegraphics[width=1\linewidth]{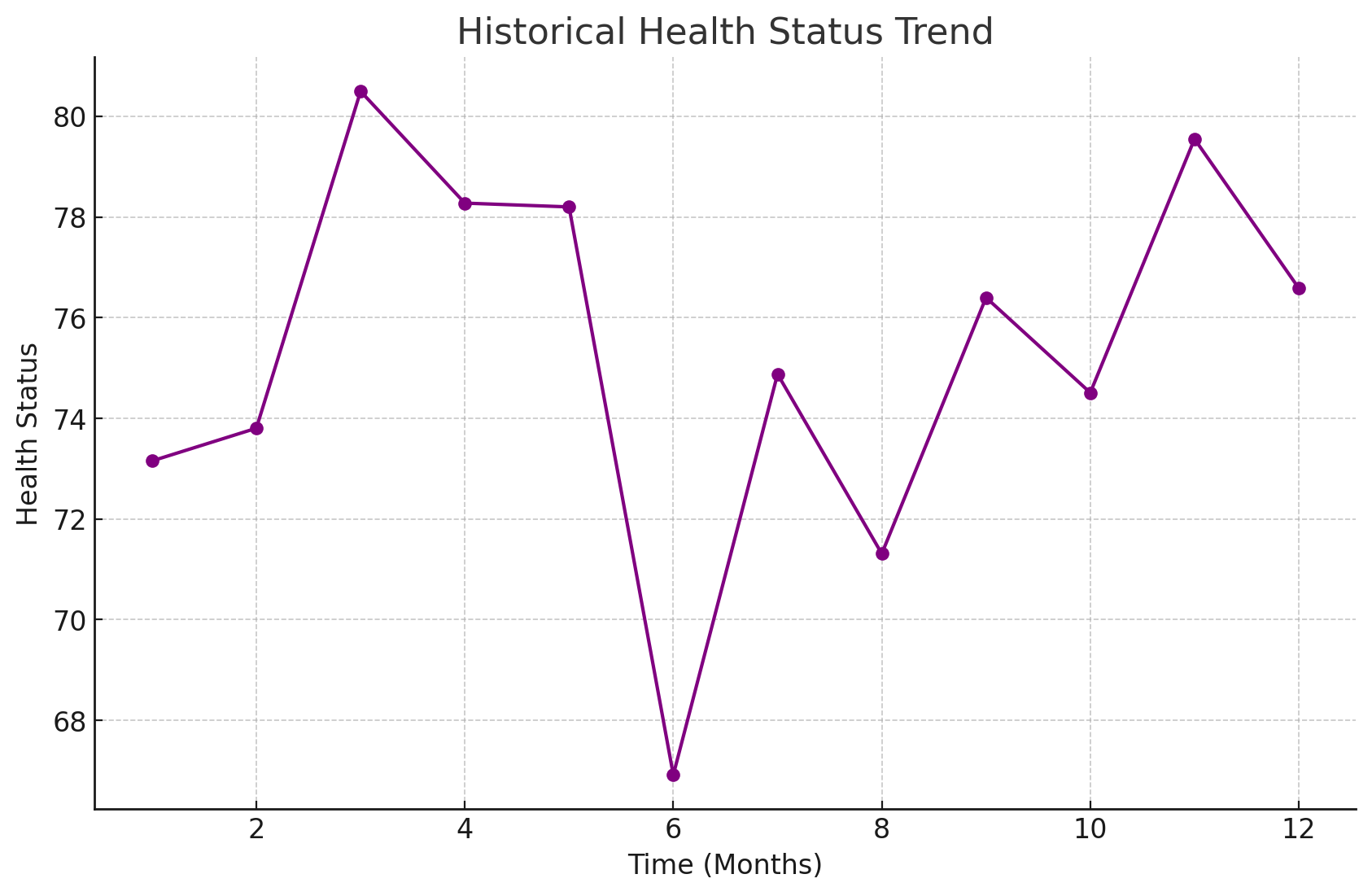}
    \caption{Health status historical trends.}
    \label{fig5}
\end{figure}

According to the survey results, the model was designed with priority given to improving the coverage and practical application effects of these functions. In addition, the service coverage of different health conditions was analyzed, as shown in Table \ref{tab2}, indicating that the current service coverage still needs to be improved in high-risk health conditions such as cardiovascular disease.

\begin{table}[htbp]
\centering
\caption{Analysis of Health Condition, Service Coverage, Mortality Rate, and Personalization Needs.}
\resizebox{\linewidth}{!}{
\begin{tabular}{|l|c|c|c|c|}
\hline
\textbf{Health Condition}  & \textbf{Population Affected (\%)} & \textbf{Service Coverage (\%)} & \textbf{Mortality Rate (\%)} & \textbf{Personalization Needs (\%)} \\ \hline
Cardiovascular             & 35                                & 65                             & 15                           & 70                                  \\ \hline
Respiratory                & 20                                & 50                             & 12                           & 60                                  \\ \hline
Diabetes                   & 15                                & 40                             & 8                            & 55                                  \\ \hline
Mobility Issues            & 30                                & 55                             & 10                           & 65                                  \\ \hline
\end{tabular}}
\label{tab2}
\end{table}

The results of these market surveys helped us clarify the key goals in the current model design and optimize resource allocation and service design through a data-driven approach.

\subsubsection{Experimental results and analysis}
The experimental results show that the proposed health behavior prediction model performs well in multiple dimensions. With the support of data fusion and missing data processing, the model demonstrates efficient prediction capabilities in a multimodal data environment. At the same time, through the nonlinear prediction and emergency detection modules, the model performs well in dealing with sudden changes in health status. Figure 1 shows the comparison between the importance of functions in market research and the current availability, reflecting the functional gaps in the current market. As shown in Figure \ref{fig1}, although real-time monitoring and prediction accuracy are of high importance, their availability in the current market is not ideal, indicating that there is a lot of room for improvement.

In the analysis of health status service coverage, as shown in Figure \ref{fig2}, there are significant differences in service coverage under different health statuses. Especially in high-risk groups such as cardiovascular disease, the service coverage still needs to be further improved to meet actual needs.

In addition, the line graph of the importance and implementation difficulty of platform functions (as shown in Figure \ref{fig3}) reveals the trade-off between the importance and implementation difficulty of different functions during system design. Figure \ref{fig3} shows that data security and interoperability are of high importance, but are also relatively difficult to implement, reminding us that we need to focus on overcoming these technical difficulties during the design process.

The health prediction results in the experiment were further verified by visual analysis. Figure \ref{fig4} shows a scatter plot of the health status predicted by the model and the actual health status. The red dotted line in Figure \ref{fig4} represents the situation where the prediction under the ideal state is consistent with the actual value. The results show that most of the prediction points are close to the dotted line, indicating that the model performs well in prediction accuracy.

Finally, Figure \ref{fig5} shows the historical trend of changes in the health status of an elderly person during the experiment. Through the time series analysis of health status, it can be found that the model can capture the changing trend of health status promptly and make predictions and adjustments at key nodes to ensure that the elderly receive timely health interventions.

\subsubsection{Extended Experiment}
To comprehensively evaluate its applicability and accuracy in different elderly populations, we designed multiple environments to test the health status of different elderly populations.

1) Urban vs. rural environment

Experimental design:
Sample selection: 500 elderly people in urban and rural areas were selected.
Data collection: Collect data on their lifestyle, medical records, environmental pollution exposure, etc.
Test items: Cardiovascular health, respiratory system function, mental health status.

Experimental results:
Urban elderly: The incidence of cardiovascular disease is higher, which may be related to high-pressure life and air pollution.
Rural elderly: Fewer respiratory diseases, but more musculoskeletal problems, which may be related to physical labor.

2) High-income vs. low-income elderly population

Experimental design:
Sample selection: According to income level, 500 high-income and low-income elderly people were selected.
Data collection: income, access to medical care, eating habits, exercise frequency.
Test items: prevalence of chronic diseases, nutritional status, mental health.

Experimental results:
High-income group: better control of chronic diseases and better mental health.
Low-income group: higher incidence of chronic diseases such as hypertension and diabetes, more common symptoms of depression and anxiety.

3) Active lifestyle vs. sedentary lifestyle

Experimental design:
Sample selection: 500 active and 500 sedentary elderly people were selected based on their daily activity levels.
Data collection: exercise habits, physical function test results, cognitive function test.
Test items: cardiopulmonary function, muscle strength, cognitive ability.

Experimental results:
Active elderly people: better cardiopulmonary function and muscle strength, better cognitive ability.
Sedentary elderly people: more prone to cardiovascular problems and cognitive decline.

4) Different dietary patterns

Experimental design:
Sample selection: According to dietary habits, the subjects were divided into balanced diet, high-fat and high-sugar diet, and vegetarian groups, with 300 elderly people in each group.
Data collection: Diet records, body mass index, blood sugar and cholesterol levels.
Test items: Metabolic health indicators, cardiovascular risk assessment.

Experimental results:
Balanced diet group: Normal weight and metabolic indicators, low cardiovascular risk.
High-fat and high-sugar group: High obesity rate, high blood sugar and cholesterol levels, increased cardiovascular risk.

5) Social activity

Experimental design:
Sample selection: According to the frequency of social activities, the elderly were divided into three groups: high, medium and low, with 400 elderly people in each group.
Data collection: Social activity records, mental health assessment, cognitive function test.
Test items: Depression and anxiety assessment, memory and attention test.

Experimental results:
High social activity: Good mental health and good cognitive function.
Low social activity: More obvious symptoms of depression and anxiety, and increased risk of cognitive decline.

6) Environmental pollution exposure

Experimental design:
Sample selection: 500 elderly people living in high-pollution and low-pollution areas were selected.
Data collection: Environmental pollution indicators, respiratory health data, long-term disease records.
Test items: Lung function test, respiratory disease screening.

Experimental results:
High-pollution areas: The incidence of respiratory diseases increased significantly and lung function decreased.
Low-pollution areas: The respiratory health status is better.

\section{Conclusion} 

In view of the multiple challenges in predicting the health behavior of the elderly, this study proposed and implemented a smart elderly care service model. The model integrates multiple advanced technologies such as multimodal data fusion, nonlinear prediction, data missing processing, emergency detection and privacy protection to form a health management platform for the elderly with high robustness and accuracy. Through market research and analysis, we clarified the key functional requirements of current smart elderly care services and verified the superior performance of the model in practical applications through experiments. The experimental results show that the model not only significantly improves the accuracy of health behavior prediction for the elderly, but also enhances the robustness and security of the system through effective data fusion and processing mechanisms. In addition, the model performs well in coping with sudden changes in health status and providing personalized services.

In future research, the model will continue to be optimized to expand its application scope and data set coverage, especially for the elderly groups in different regions and cultural backgrounds. Further integration of artificial intelligence and deep learning technologies can improve the adaptability and intelligence level of the system. At the same time, with the increasing demand for smart elderly care services, the model needs to be optimized in terms of large-scale deployment and real-time response to meet a wider range of market needs. In summary, this study provides important technical support for the development of smart elderly care services and lays a solid foundation for future research in related fields.

\section*{Conflict of Interest Statement}

The authors declare that the research was conducted in the absence of any commercial or financial relationships that could be construed as a potential conflict of interest.

\section*{Author Contributions}
\textbf{Qian Guo:} Data Management, Methodology and Writing – original draft. \textbf{Peiyuan Chen:} Software, Supervision, Writing – review \& editing.

\section*{Acknowledgments}
The research was supported by the Youth Project of Social Science Foundation of Anhui Province under grant Nos. AHSKQ2022D069..

\bibliographystyle{Frontiers-Vancouver} 
\bibliography{test}

%%% If you don't add the figures in the LaTeX files, please upload them when submitting the article.
%%% Frontiers will add the figures at the end of the provisional pdf automatically
%%% The use of LaTeX coding to draw Diagrams/Figures/Structures should be avoided. They should be external callouts including graphics.

\end{document}